# Evaluating the Performance of AI Text Detectors, Few-Shot and Chain-of-Thought Prompting Using DeepSeek Generated Text

Hulayyil Alshammari, Praveen Rao
*Department of Electrical Engineering & Computer Science*
*The University of Missouri, Columbia, USA*
hamkx@umsystem.edu, praveen.rao@missouri.edu

*Abstract*— Large language models (LLMs) have rapidly transformed the creation of written materials. LLMs have led to questions about writing integrity, thereby driving the creation of artificial intelligence (AI) detection technologies. Adversarial attacks, such as standard and humanized paraphrasing, inhibit detectors' ability to detect machine-generated text. Previous studies have mainly focused on ChatGPT and other well-known LLMs and have shown varying accuracy across detectors. However, there is a clear gap in the literature about DeepSeek, a recently published LLM. Therefore, in this work, we investigate whether six generally accessible AI detection tools, namely, AI Text Classifier, Content Detector AI, Copyleaks, QuillBot, GPT-2, and GPTZero, can consistently recognize text generated by DeepSeek. The detectors were exposed to the aforementioned adversarial attacks. We also considered DeepSeek as a detector by performing few-shot prompting and Chain-of-thought reasoning (CoT) for classifying AI and human-written text. We collected 49 human-authored question-answer pairs from prior to the LLM era and generated matching responses using DeepSeek-v3, hence producing 49 AI-generated samples for the evaluation. Next, we utilized adversarial techniques such as paraphrasing and humanizing to add 196 additional adversarial attack samples to the dataset. These reinterpreted samples were used to challenge the detectors' tolerance or resistance to adversarial attacks and assess their impact on detection accuracy. While QuillBot and Copyleaks achieved near-perfect recognition rates for original and paraphrased DeepSeek text, other tools—particularly AI Text Classifier and GPT-2— struggled to show consistent detection performance. Paraphrasing and humanizing DeepThink-generated text reduced the accuracy of GPTZero, Copyleaks, and QuillBot. The most effective adversarial attack was text humanization, which reduced the accuracy to 71% for Copyleaks, 58% for QuillBot, and 52% for GPTZero. The few-shot and CoT techniques of DeepSeek yielded significantly higher accuracy scores. The highest score achieved by few-shot prompting was for the five-shot setting, which misclassified only one sample out of 49 test samples (AI Recall 96%, Human Recall 100%). The CoT prompting also presented a high accuracy result (AI Recall of 92.6% and Human Recall of 90%). Our findings highlight the need for continuous improvements in AI content identification as LLM architectures continue to evolve. We emphasize that AI detectors should be carefully selected based on users' requirements in research, professional, and educational environments.

*Keywords*— LLM, Academic Integrity, Content Detection, Deepseek, few-shot, Chain-of-Thought, Machine-Generated TexIntroduction (Heading 1)

## I. Introduction

In fields such as higher and secondary education, students experience an intensive and organized learning process, which may involve exams, quizzes, assignments, projects, etc. These activities provide the necessary level of knowledge and skills that every student can gain through iteration and practice. Following these educational processes is a foundational factor in measuring students' performance [1][2]. However, artificial intelligence (AI) is drastically changing educational processes [3] [4]. While education is the most affected field, AI, or large language model (LLM)-generated content, has also shifted how other fields function [5] [6]. In this study, the educational sector was selected as an example and illustration of the drawbacks of LLM misuse.

AI is becoming a significant factor in education [4]. Through smart tutoring resources and services, it can improve educational techniques and learning experiences [7]. However, like everything in life, this has advantages and disadvantages. One main downside of AI in general and LLM in particular is the potential breach of academic integrity [8].

A well-known example of an LLM is ChatGPT. ChatGPT is a chatbot that was built on an LLM model [5]. LLM services are considered one of the major threats to academic integrity since they can be misused [9] [10]. With tools such as ChatGPT available online, students can spend less time and effort learning due to the tremendous support these tools offer [11] [12]. Students may use LLM to solve quizzes and exams, assignments, write essays, etc. This use may damage or affect the learning process that students should be involved in [11], because the learning process is a chain of cycles represented by activities or tasks that every student should go through to achieve the maximum benefit of his or her time in an educational institute. With LLM completing most of the work, this chain is no longer effective.

Since its release in late 2022, ChatGPT has become the most famous and used LLM chatbot[13]. However, a new LLM model, Deepseek, was released at the beginning of 2025 and offered a free and efficient LLM chatbot [14]. As a result of the popularity of LLM tools, new types of LLM content detection and verification tools or services have emerged [15] [13]. LLM content detectors are used to identify machine-generated content from human content. Their role is essential in ensuring integrity and transparency in fields such as education, social media, and news. AI content detectors were tested with text mainly from ChatGPT and other LLMs, such as Microsoft Copilot or Google Gemini [15] [11]. However, will the AI content detection tools be able to detect AI content generated by the newcomer, Deepseek? There are several LLM content detection tools, and their accuracy scores vary [16]. Choosing an AI text detector that suits all LLM-generated content is essential. On compiling this paper in March 2025, and based on our search in Google Scholar, we did not find research that involved testing LLM/AI content detectors, few-shot and Chain of Thoughts (COT) on Deepseek-generated text which makes this work timely.

Although LLM text detectors are available online, choosing one can be a nuanced decision. The selection is determined by the associated utilization cost, the detector's accepted maximum and minimum token limit, and, most importantly, by the detector's accuracy. Deepseek's rising prominence highlights the importance of evaluating the accuracy of AI detectors on content it generates. In this study, several LLM text detectors were tested using Deepseek-generated content, and the reliability and accuracy scores of the detectors were identified. The study tested six detection tools' abilities with human text, Deepseek text, and Deepseek-paraphrased text. Our primary inquiry is whether detectors can identify content generated by Deepseek and to what extent their accuracy can be guaranteed. Furthermore, an additional investigation was performed to assess the text classification performance in Deepseek using few-shot and COT techniques. Key contributions of this work are the following:

- **Systematic evaluation of Deepseek-generated text:** By evaluating six AI-text detectors (AI Text Classifier, Content Detector AI, Copyleaks, QuillBot, GPT-2, and GPTZero) on content generated by the recently published Deepseek model, the study fills a glaring gap in the literature.
- **Balanced human vs. LLM-generated corpus creation:** The study generated a controlled testbed consisting of 49 human-written Q&A pairs and 49 corresponding Deepseek answers for a thorough assessment. Then, paraphrased and DeepThink text versions were produced.
- **Analysis of detector robustness to paraphrasing:** The study determines which systems retain accuracy under content alteration and which do not by contrasting detector performance on the original Deepseek outputs with both Deepseek paraphrased and DeepThink-paraphrased versions.
- **Investigation of prompting techniques to boost detection:** This work assesses few-shot and chain-of-thought (CoT) prompting applied to Deepseek, going beyond commercial detectors, and shows that minimal in-prompt examples or explicit reasoning steps significantly increase AI/human classification accuracy.

## II. RELATED WORK

Detecting Machine Generated Text (MGT) has been an active topic of discussion ever since the LLMs based chatbots became popular and widely used. There have been several innovative detection techniques such as those in [17] and [18]. However, due to the continuous improvements and releases of new LLM models, the detection process becoming increasingly difficult since the LLM is mimicking the human writing style more efficiently with every new release of an LLM model.
.
Several research papers examine AI content detectors with LLM-generated content. Halaweh and El Refae [15] investigate AI content detection software performance with ChatGPT-generated content. The results show that the AI detection software used in this investigation could not detect AI text after being paraphrased several times by ChatGPT. This study presents important results for AI content detector users, such as teachers, university faculty, and researchers. Educators and academics need not establish a certain threshold or percentage to ascertain what constitutes appropriate AI-generated text. Establishing such a threshold may be deceptive, given the existing restrictions in these instruments' algorithms.

Orenstrakh, Karnalim, Suárez, and Liut [11] examine eight LLM detectors' performance and accuracy. The work was conducted in computer education, where they collected 128 submissions that belonged to computer science students in three courses. The authors made sure the submissions were compiled before the ChatGPT release. They created 40 ChatGPT submissions for the evaluation process. Their main goal was to measure accuracy, false positives, and resilience. Based on their results, Copyleaks was the best detection tool for accuracy, GPTKit presented the best results for reducing false positives, and GLTR was the most resilient or flexible detector. Moreover, based on their observations and tests during the study, the eight LLM content detectors' accuracy decreased when tested with code, languages other than English, and paraphrased text. Both studies used several methods, tools, and techniques to evaluate detection software performance, with ChatGPT content the only AI content used. However, in this paper, we only assess Deepseek-generated content and the performance of six detectors.

## III. METHODOLOGY

The data was collected from previously posted questions, primarily sourced from Quora[19] and Academia[20], dating back several years **(from 2011 to 2021)**. The original data consist of human-written questions and answers from several domains, as presented in Table I. Subsequently, a sample of 49 responses was generated using Deepseek for the original 49 questions that humans answered. The Deepseek responses were collected from the chatbot (Deepseek-V3), meaning we asked Deepseek all the questions and saved its responses as our AI-generated text data. Following this, 49 responses generated from Deepseek were paraphrased using Quillbot paraphrasing tool [19]. Moreover, another data sample was generated using the DeepThink feature. All samples, including human-written, Deepseek-original, Deepseek-paraphrased, DeepThink-text, and DeepThink-paraphrased, were used in the detector's evaluation process.

TABLE I PRESENTS THE ORIGNAL QUESTIONS PER DOMAIN

| Academia | Business | History | Life | Entertainment | Others |
|---|---|---|---|---|---|
| 22 | 6 | 7 | 5 | 4 | 5 |

TABLE II NUMBER OF TESTING SAMPLES PER CATEGORY

| Sample Category | Number of Samples |
|---|---|
| Human-text | 49 testing samples |
| Deepseek-text | 49 testing sample |
| Deepseek-Paraphrased text | 49 testing sample |
| Deepthink-text | 49 testing sample |
| Deepthink-Paraphrased text (Standard) | 49 testing sample |
| Deepthink-Paraphrased (Humanized) | 49 testing sample |

Table II represent the collected data, which contain 294 testing samples consisting of answers. Of these, 49 were human-written, 49 were DeepSeek-generated, 49 were DeepSeek-paraphrased, and the remaining 147 were DeepThink's responses (including DeepThink generated and paraphrased). The human samples were collected from various questions covering domains such as general life, history, economy, chemistry, physics, and technology. We recorded all questions and answers and included a link to every question that will be available in an Excel sheet for reference. Every human and Deepseek answer was evaluated using six detectors. Each tool operates using specific methodologies and techniques in the results generation process. The detectors present results as percentages, indicating human probability or AI probability; labels, such as real or fake; metrics, such as mixed text percentages; or solely report the probability that the text was AI-generated. We were forced to delete portions of our data samples on several occasions due to text or token limitations in some of the detectors. We also encountered challenges with some detectors not providing any results for certain samples. This issue occurred most frequently with the AI Text Classifier.

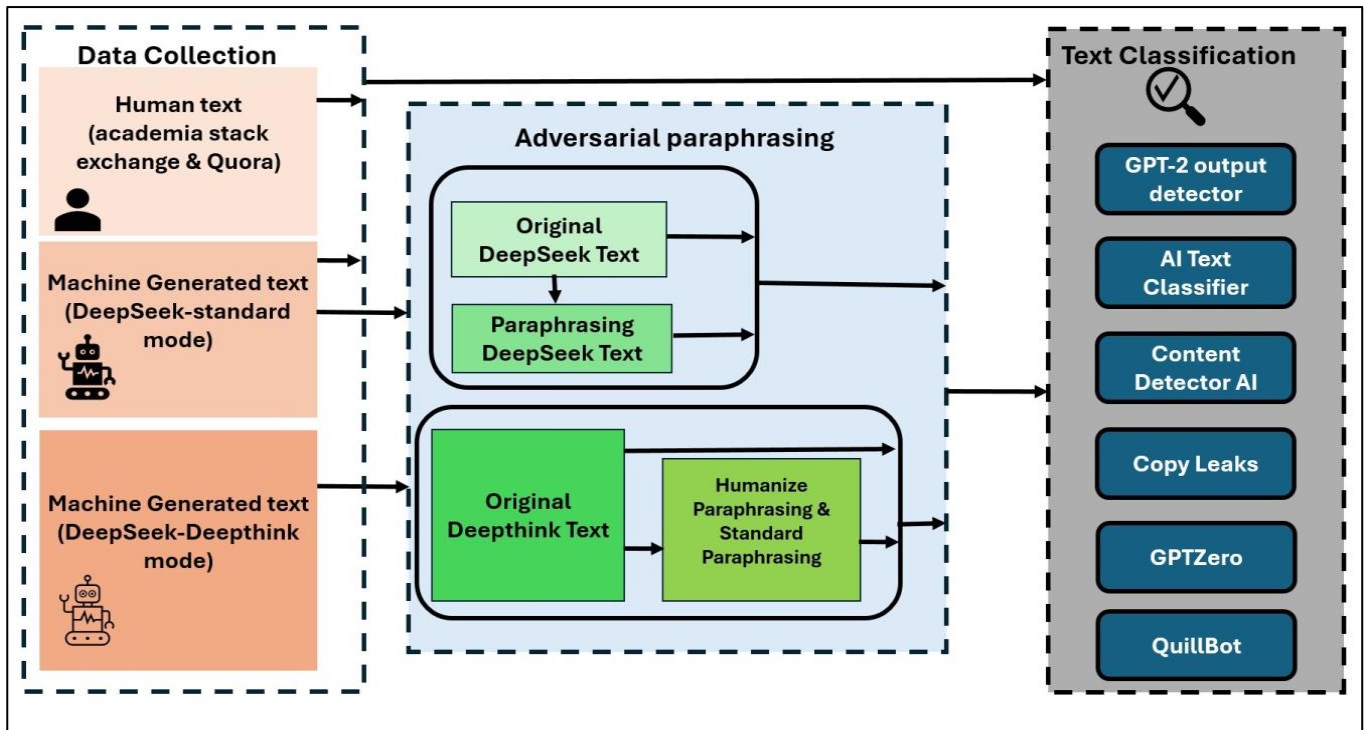

**Fig. 1.** *illustrates the end-to-end process of collecting human Q&A data, generating answers with Deepseek, and testing multiple detectors on human, Deepseek, Deephink, and paraphrased text.*

The detectors' assessment process involved four main phases. The first phase was testing the detectors with 49 human text samples and calculating every detector's average accuracy (Table III). The second phase involved measuring the AI detectors' accuracy with Deepseek-generated text (Table III). The third phase tested the detectors' average accuracy with Deepseek paraphrased text (Table IV). Finally, the 4th phase included testing DeepThink- paraphrased and original content (Table V). The "AI Text Classifier" was excluded from the 4th phase because it did not respond to the paraphrased text, meaning it did not provide any results when Deepseek-paraphrased content was entered into its text box. Detectors with low accuracy scores in the first, second, and third phases were excluded from phase 4. The study also conducted a separate assessment of few-shot and chain-of-thought learning techniques for AI/human text classification.

## IV. AI CONTENT DETECTION TOOLS

A method was needed to determine whether a human or a machine wrote a text. Therefore, AI-detection tools were developed. Numerous LLM text detection tools are available, and, for this paper, we conducted research using six detectors. Since free tools are more widely used, we attempted to avoid paid detection services. However, we ultimately used a combination of free and paid detectors. The choice of our selection of detectors used in this research is based on popularity and tools that are mentioned in the Texas Tech University Libraries [21] and by Orenstrakh, Karnalim, Suárez and Liut [11]. **AI Text Classifier [22]:** Free online AI content detector that shows the percentage of human and AI content. **Content Detector AI [23]:** Content Detector AI was developed with tasks that allow it to detect whether a text is generated by AI. It produces a probability score of whether a text is generated by AI or written by a human. More than a million bloggers and academics have used this tool [23]. **Copyleaks [24]:** A powerful tool for AI text detection used by regular users and professionals, such as schoolteachers and university faculty, to ensure that a text sample is written by a human, not AI or LLM. Copyleaks can detect content generated from several LLMs, such as, but not limited to, ChatGPT and Jasper. The tool reported 99.12% accuracy in content detection tasks [25]. **QuillBot [26]:** This detector, known as a paraphrase tool, also provides AI content detection services and features. The tool also provides other services, such as grammar checks, citation generation, and summarizations [27]. **GPT-2 [28]:** Developed by OpenAI, it offers detection services to discern AI text from human text. GPT-2 is an LLM that was fine-tuned based. With an accuracy of 88% at 124 million parameters and 74% at 1.5 billion parameters, the GPT-2 detector provides a likelihood that an input text is authentic [29]. **GPTZero [30]:** A classification model that was developed by Edward Tian to help workers in the education field detect AI-generated content. The tool was trained on a variety of human and machine texts. The language used in the data was English [25].

## V. RESULTS AND DISCUSSION

### 1. Detectors Evaluation

#### A. Accuracy of AI Content Detectors

Tables III, IV, and V represent the detectors evaluation results. When we test a detector with a text sample it will give back a percentage of AI text which is a value between 0 and 100. Every text sample has an AI text percentage based on the value provided by the detectors. The **average accuracy score** presented in Table III, IV and V is the arithmetic mean for the AI text percentages giving by the detectors. The accuracy score for human/AI text detection presented in Table III was calculated in the same method. The final accuracy score for all detector was calculated using the **arithmetic mean (average),** which is represented in the following formula:

$$\text{Average AI Detection} = \frac{\sum X_i}{N},$$

where:
- $X_i$ = Each AI detection percentage in the dataset
- N = Total number of values (samples)

**To illustrate,** if we are evaluating Detector X using 4 text samples, with the percentages of AI text correctly identified being 80%, 75%, 66%, and 70%. The average accuracy score of Detector X would be the **arithmetic mean** of these values. This method was used when evaluation the detectors accuracy with Human text, AI text, AI paraphrased text and DeepThink text. More evaluation metrics were used to calculate the accuracy of the detectors with AI text, such as **Recall metrics (at the 50% threshold),** which is used to evaluate the detectors performance with AI text, AI paraphrased text and DeepThink text as shown in Fig 2, 3 and 4. In this metric any sample is considered AI text if it contains more than 50% of AI text. **True Positive (TP)** refer to the correctly identified samples while **False Negative (FN)** refers to the incorrectly identified samples.

$$\text{Recall} = \frac{TP}{TP+TN}$$

#### B. Human Content Detection

Table III displays the results recorded for the average accuracy score of every detection tool when tested with a human text sample. The results showed that QuillBot and Copyleaks achieved the highest scores of 100% in human-written text detection, followed by AI Text Classifier at 98.4%, GPTZero at 98.2%, and Content Detector AI at 55.2%. The detection tools were tested with human samples for a total of 294 attempts. Based on the high accuracy and free membership, QuillBot appears to be the ideal option for human text detection.

TABLE III. DETECTOR'S AVERAGE ACCURACY IN DETECTING HUMAN AND AI TEXT

| Detector name | Number of Samples | Average accuracy score-AI | Average accuracy score-Human |
|---|---|---|---|
| AI Text Classifier | 49 sample | 3.24% | 98.14% |
| Content Detector AI | 49 sample | 67.27% | 55.2% |
| Copyleaks | 49 sample | **98.8%** | **100%** |
| QuillBot | 49 sample | 98.4% | **100%** |
| GPT-2 | 49 sample | 8.15% | 93.73% |
| GPTZero | 49 sample | 100% | 98.2% |

#### C. Deepseek Content Detection

Table III provides statistics for the testing process that was conducted using Deepseek-generated data. The outcomes showed that GPTZero had the highest accuracy score among the detection tools, with 100%, followed by CopyLeaks with 98.8%, and QuillBot with 98.4%. The performance of the remaining detectors was unsatisfactory, as their accuracy score was considerably lower compared with Quillbot, GPTZero, and Copyleaks. The detector with the lowest score was AI Text Classifier, with an average score of 3.24%, followed by GPT-2, scoring 8.15%, followed by Content Detector AI, with an average score of 67.27%. The most pronounced shift in performance was with AI Text Classifier, which achieved an accuracy score of 98.14% in human sample detection. In contrast, its performance declined to 3.24% when evaluated with Deepseek-generated text. The remaining detectors' accuracy was roughly similar when testing with a human or AI sample, except for GPT-2 and AI Text Classifier. Figure 4 represents the recall results of this phase with a 50% threshold.

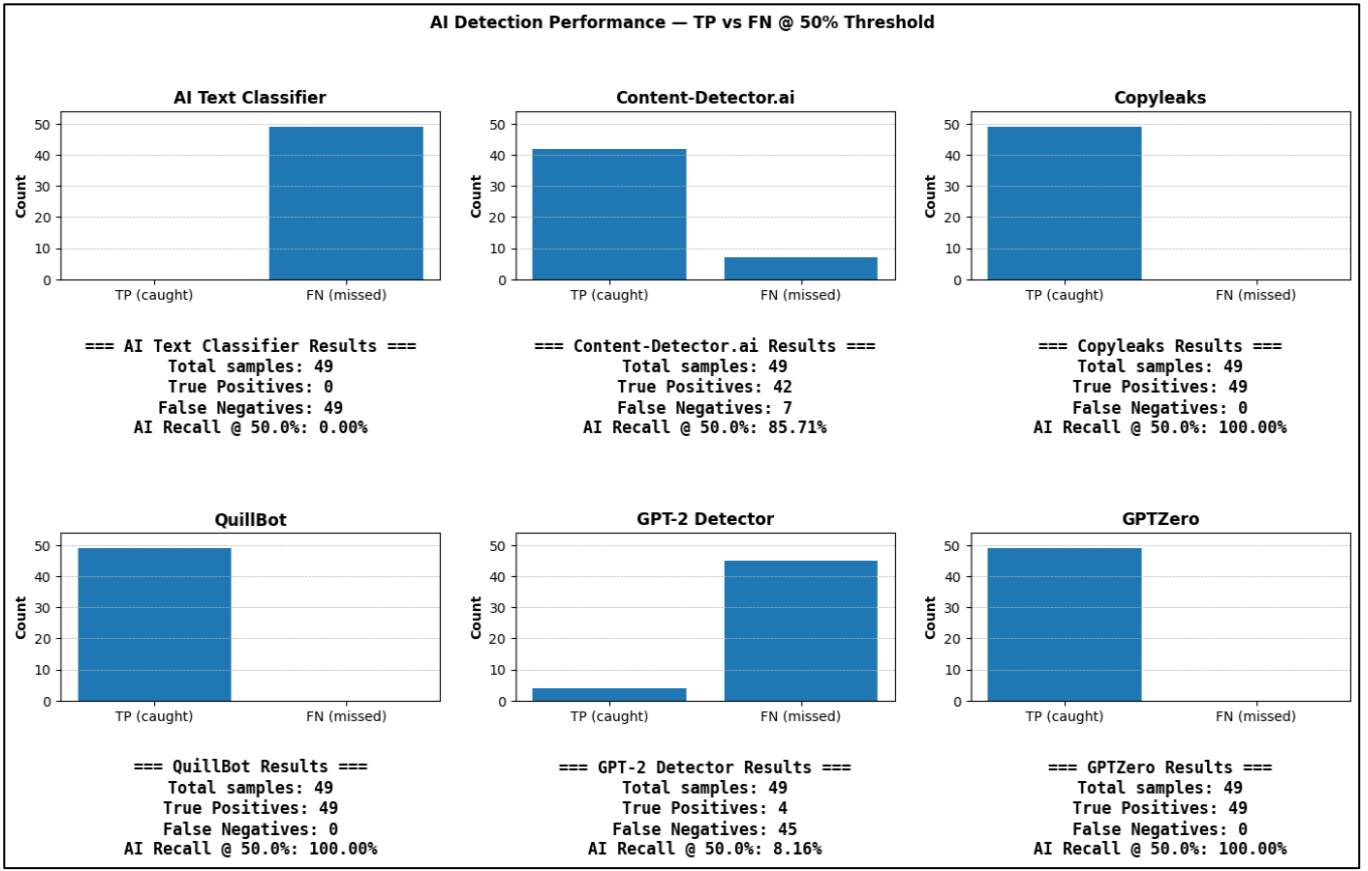

Fig. 2. *Measuring the recall of AI text detectors at a 50% decision threshold for Deepseek-text.*

## D. Paraphrased Content Detection

Following the evaluation process of AI detectors with Deepseek and human content, we investigated further, testing, and challenging the tools with more complex data samples. We paraphrased 49 responses generated by Deepseek and tested the detection tools using the paraphrased data. With the paraphrased text testing, QuillBot scored the highest accuracy at 95.49%, followed by Copyleaks at 93.54%, GPTZero at 92.61%, Content Detector AI at 59.25%, and GPT-2 with 3.65%. Table IV presents the results of this phase, summarizing each detector's performance across all levels of testing. Figure 3 represents the recalls for this phase, while Figure 2 provides a visual representation of phases 2, illustrating each detector's relative effectiveness under varying conditions across the stages of testing.

TABLE IV. DETECTORS' ACCURACY IN DETECTING AI PARAPHRASED TEXT.

| Detectors | Number of attempts | Sample type | Average accuracy score |
|---|---|---|---|
| Content Detector AI | 49 attempts | AI | 59.25% |
| Copyleaks | 49 attempts | AI | 93.54% |
| QuillBot | 49 attempts | AI | **95.49%** |
| GPT-2 | 49 attempts | AI | 3.65% |
| GPTZero | 49 attempts | AI | 92.61% |

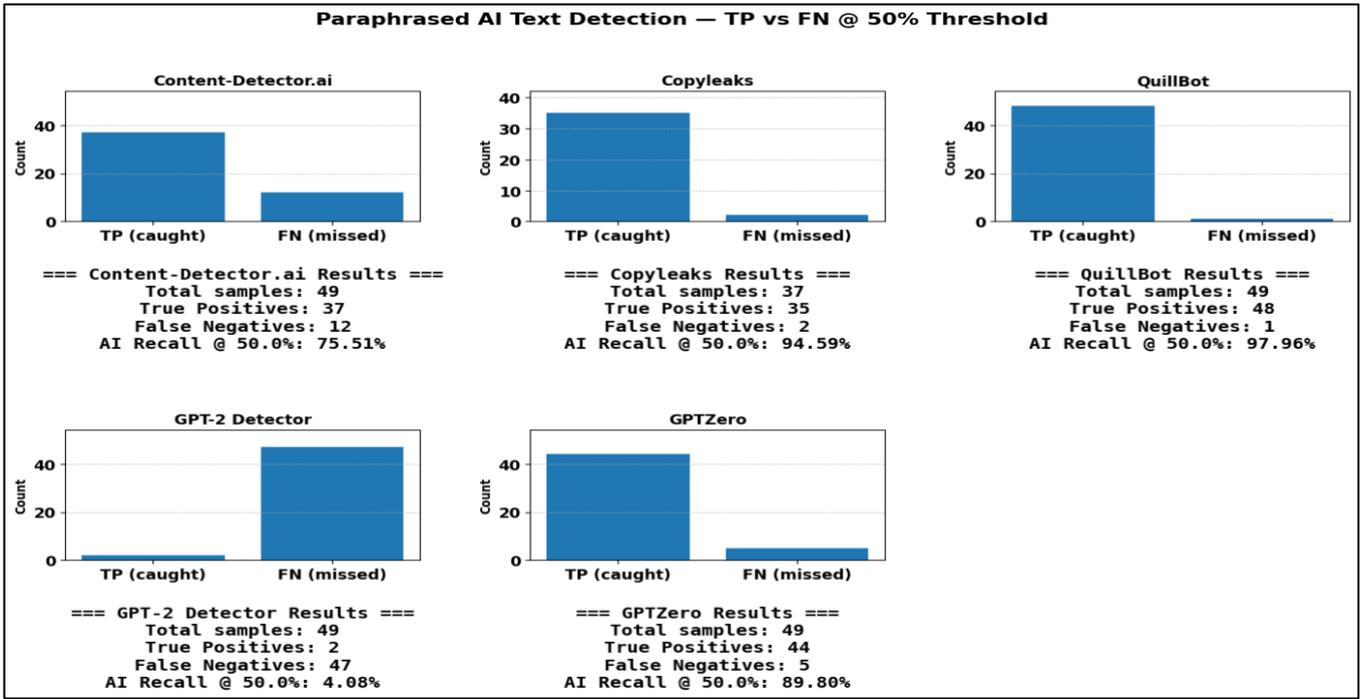

Fig. 3. *Measuring the Recall of AI text detectors at a 50% decision threshold for the DeeSpseek-paraphrased text phase, showing the performance using the Recall matric .*

### E. DeepThink Content Detection

After completing phases 1, 2, and 3 of the evaluation process, we selected the detectors with the highest accuracy scores for our final phase of detectors evaluation process. We activated the DeepThink feature in Deepseek to generate text to be used in this phase. Subsequently, two distinct methodologies were employed to paraphrase the generated text: a humanized phrasing mode and a standard phrasing mode. Forty-nine text samples were tested using three detecors GPTZero, CopyLeaks, and QuillBot. CopyLeaks attained an accuracy score of 99.7% in the DeepThink-Text testing, followed by QuillBot with 95.4% and GPTZero with 94.1%. With a score of 84.1.90%, QuillBot performed the highest score in the Deepthink- paraphrased (Standard Mode) evaluation, followed by GPTZero at 82.3% and CopyLeaks at 81.2%. Recording 71%, CopyLeaks outperformed other detectors in the DeepThink-paraphrased-text (humanize mode) stage; QuellBot scored 58%; and GPTZero had the lowest score at 52%. The results were also calculated using the recall with a 50% threshold as shown in Figure 4. These findings offer a thorough contrast of every tool's performance in producing and paraphrasing text, applying both conventional and humanized approaches.

This study was conducted in March 2025, and the accuracy of the detectors stated in this research was recorded in the same month. Based on our observations and the results during this study, the following detectors need updates: AI Text Classifier, Content-Detector AI, and GPT-2. QuillBot could be more efficient if it offered APIs in addition to its free services. A common issue with most of the detectors is the minimum text limit allowed. Based o the research, the minimum text required for a detection process for QuillBot is 80 words, for Copyleaks, it is 350 characters, and for GPTZero, it is 250 characters. In this work, two out of six detectors scored extremely low in terms of accuracy, and one scored around 60% (Tables III and IV). One main risk we noted is the false positives issue, when a detector recognizes or predicts a human text as a machine text. QuillBot and Copyleaks did not present false positive cases. The detector that showed the most false-positives was Content Detector AI, followed by GPT-2 and GPTZero. Paraphrasing DeepThink content significantly drops the accuracy scores as presented in Table V. Overall, the findings indicate that not all the current online detection systems are 100% reliable. Although Copyleaks, QuillBot, and GPTZero demonstrated high accuracy across all testing phases, their performance declined notably when evaluated with DeepThink-paraphrased text.

TABLE V. THE DETECTORS' ACCURACY IN DETECTING DEEPTHINK MODE TEXT.

| DeepThink-text | | DeepThink-Paraphrasing (Standard mode) | | DeepThink-Paraphrasing (Humanize mode) | |
|---|---|---|---|---|---|
| Detector | Score | Detector | Score | Detector | Score |
| GPTZero | 94.1% | GPTZero | 82.3% | GPTZero | 52% |
| Copyleask | 99.7% | Copyleask | 81.2% | Copyleask | **71%** |
| QuillBot | 95.4% | QuillBot | 84.1.9% | QuillBot | 58% |

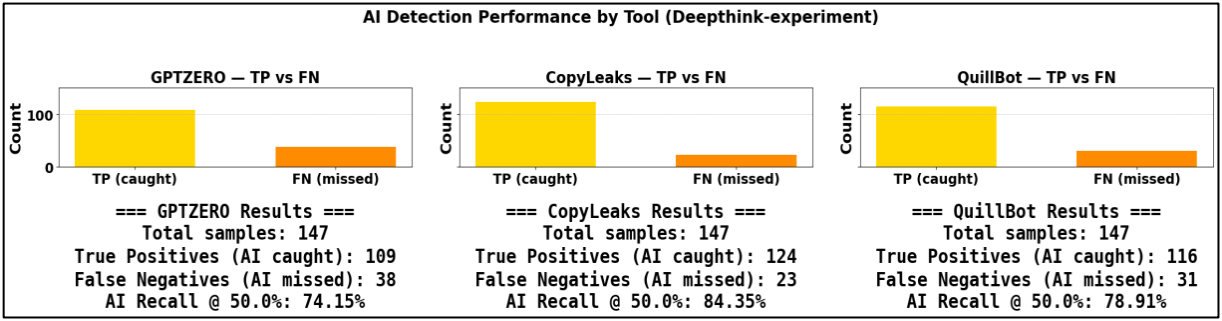

Fig. 4. *Measuring the recall of AI text detectors at a 50% decision threshold for DeepThink-text, DeepThink-paraphrased-text (standard mode) and DeepThink-paraphrased-text (Humanize mode)*

## 2. COT and Few-shot Classifier Accuracy

As further investigation in the AI/human text classifying techniques and tools, an evaluation was conducted on DeepSeek's ability to distinguish between AI-generated and human-written text using both few-shot prompting and Chain-of-Thought (CoT) techniques. Beginning with a zero-shot baseline and progressing through one- to five-shot demonstrations, we then apply a CoT prompting to elucidate the model's reasoning process further. Across all settings—0-shot, one-shot, two-shot, three-shot, 4-shot, 5-shot, and CoT. This comprehensive framework allows us to compare how incremental in-prompt examples and explicit step-by-step reasoning impact Deepseek's sensitivity to AI-generated content and its precision in preserving authentic human prose. A confusion representation matrix and Recall matric were used to evaluate the accuracy of CoT and few-shot method in identifying AI-generated text.

### A. Evaluating Zero and One-Shot Classifier Accuracy

The classifier performs well on human text in the zero-shot setting, correctly identifying 22 out of 24 passages (≈91.7% specificity), but only flags 18 out of 25 AI samples (≈72% sensitivity), resulting in an overall accuracy of roughly 81.6%. A single in-prompt example increases accuracy on real writing to about 96% (23/24) but lowers AI recall to about 52% (13/25), lowering overall accuracy to about 73%. The need for richer few-shot or CoT strategies is thus highlighted by the fact that, although one-shot prompting improves the model's discrimination of human prose, it is unable to expand its detection of diverse AI-generated styles.

### B. Evaluating Two and Three-Shot Classifier Accuracy

Both the two-shot and three-shot classifiers achieved flawless detection on our dataset. In each case, the model correctly labeled all 24 human-written passages (100% specificity) and all 25 AI-generated samples (100% sensitivity), yielding an overall accuracy of 100% with zero false positives or negatives. Providing just two in-prompt examples sufficed to fully align the model's decision boundaries with the classification task, and adding a third exemplar maintained this perfect performance, underscoring the power of a minimal few-shot setup to capture the stylistic distinctions between human and machine-generated text.

### C. Evaluating Four and Five-Shot Classifier Accuracy

The classifier maintains perfect specifically with four in prompt examples, correctly labeling all 24 human-written passages. At the same time, its sensitivity to AI content increases to 96% (24/25), resulting in an overall accuracy of about 98.4%. The same results are obtained by introducing a fifth exemplar: 98.4% accuracy, 96% sensitivity, and 100% specificity, with one machine-generated sample remaining incorrectly classified. The findings showed that more shots provide diminishing returns beyond a small number of carefully selected examples. That careful exemplar selection, not just quantity, is essential to fully capturing the stylistic variance of AI-generated text. Figure 5 provides a confusion matrix representation for Five-shot prompting accuracy.

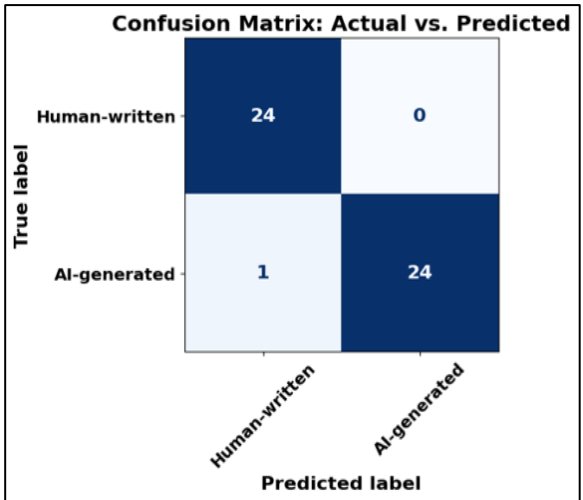

Fig. 5. Confusion matrix representations for Five-shot prompting accuracy.

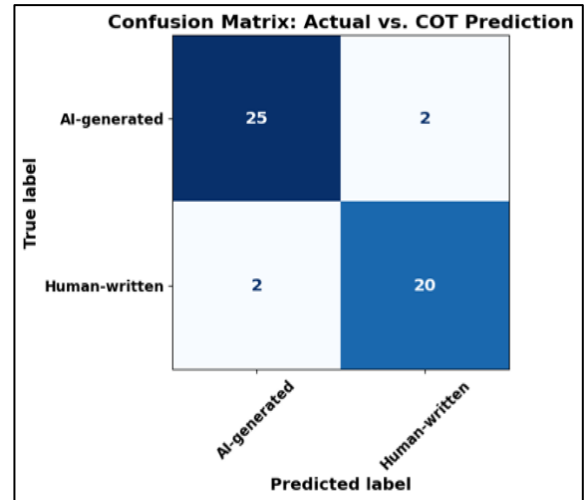

Fig. 6. Confusion matrix representations for CoT prompting accuracy results in the evaluated with 49 AI/Human samples.

### B. Evaluating Chain of Thought Techniques' Accuracy

Detection of machine-generated text based on linguistic features has been mentioned and implemented multiple times in the literature [18], [31] [32], [33]. The creation of a Chain-of-Thought (CoT) for the purpose of deepfake-textual detection relies on defining a set of criteria that distinguish human-written text from machine-generated text. Based on linguistic features discussed in the literature, along with analysis and observation of our dataset, we were able to identify a set of features to be used as criteria in the CoT. The linguistic and stylistic features used in the CoT are: (1) Generic vs. Specific Content; (2) Emotional Authenticity); (3) Structural Perfection vs. Imperfections; (4) Neutral/Impersonal vs. Personal Voice; (5) Narrative Flow & Creative Juxtapositions. Table VI is a clarification for one of the criteria used in the CoT (Neutral/Impersonal vs. Personal Voice).

TABLE VI. REASONING CRITERION AND CHAIN-OF-THOUGHT PROMPT FOR TONE IN HUMAN VS. AI TEXT CLASSIFICATION

| Reasoning Criteria: |
| --- |
| 4. Tone: Neutral/Impersonal vs. Personal Voice<br>**AI-style**: No first-person ("I," "we"), little to no subjective phrasing.<br>**Human-style**: First-person pronouns, personal anecdotes, soft qualifiers ("I felt," "in my experience"). |
| **Chain of Thought:** |
| 4. Tone: Does it use "I," anecdotes, or maintain an impersonal distance? |

The CoT classifier showed a good mix between preserving original writing and spotting AI output. It yields an AI recall of roughly 92.6% by correctly identifying 25 AI-generated passages while just misclassifying 2 as human. Conversely, for a human specificity of about **90.9%,** it correctly labels 20 of the 22 human-written texts and faults just 2 for machine-generated, as shown in Figure 6. With an overall accuracy close to 91.8%, the CoT method showed that explicit, step-by-step thinking in the prompt can significantly increase both sensitivity to AI patterns and protection of authentic language.

This investigation process revealed that, in artificial intelligence rather than human text classification tasks, both few-shot prompting and Chain-of-Thought (CoT) approaches show rather high performance. Apart from the zero- and one-shot conditions, few-shot techniques attain almost perfect accuracy with as few as two in-prompt examples; this high degree of performance holds as the number of shots rises. Although its accuracy falls somewhat short of the ideal scores obtained by the few-shot configurations, CoT prompting also produces strong classification results.

### VI. LIMITATION

Through this research, we encountered two main challenges: dataset limitations and commercial detector subscriptions. Due to the recent release of DeepSeek, we were unable to find a reliable dataset to use in this study. This is the primary reason for the limited size of our dataset. Although we built our dataset based on 49 questions, the total amount of generated data is far more than 49 samples. The dataset includes guaranteed human-written text, AI-paraphrased text, humanized text, and DeepThink-generated text. Another limitation is related to the use of the commercial detectors. GPTZero and Copyleaks require subscriptions; however, even with the subscription, we reached the token limits very fast—before finishing 30 testing samples.

### VII. CONCLUSION

This work evaluated AI detectors using Deepseek-generated text and performed complementary analysis using Deepseek's few-shot and CoT performance for classifying AI/Human text. New data was collected from several sources and created by Deepseek. The research showed that 50% of the detectors used in the study presented low accuracy scores, while the other 50% presented high reliability performance. Adversarial paraphrasing can limit the accuracy of the AI detectors. The study also revealed a promising performance for few-shot and CoT in the AI/Human text classification abilities. Based on our research before starting this study, challenging AI detectors with a Deepseek-generated text has not been conducted which make this work a novel and timely. Moreover, Deepseek's few-shot, and CoT prompts have not been assessed for similar scenarios.